\PassOptionsToPackage{prologue,dvipsnames}{xcolor}

\documentclass[10pt,twocolumn,letterpaper]{article}

\usepackage[pagenumbers]{cvpr} 

\usepackage{amsmath}
\usepackage{amsthm}
\usepackage{derivative}
\usepackage{tikz}
\usetikzlibrary{cd}
\usepackage{cuted}
\usepackage{url}

\usepackage[dvipsnames]{xcolor}

\definecolor{cvprblue}{rgb}{0.21,0.49,0.74}
\usepackage[pagebackref,breaklinks,colorlinks,citecolor=cvprblue]{hyperref}
\usepackage{bm}
\usepackage{svg}
\usepackage{makecell}

\def\paperID{6078} 
\def\confName{CVPR}
\def\confYear{2024}

\title{MinBackProp~-- Backpropagating through Minimal Solvers}

\author{
\parbox{0.4\textwidth}{\centering
Diana Sungatullina$^{1,2}$\\
{\tt\small diana.sungatullina@cvut.cz}
}
\hspace{0.05\textwidth}
\parbox{0.4\textwidth}{\centering
Tomas Pajdla$^1$\\
{\tt\small pajdla@cvut.cz} \\
}\vspace{0.5cm} \\
$^1$ Czech Institute of Informatics, Robotics and Cybernetics, Czech Technical University in Prague \\
$^2$ Faculty of Electrical Engineering, Czech Technical University in Prague\\
}

\newlist{todolist}{itemize}{2}
\setlist[todolist]{label=$\square$}
\usepackage{pifont}

\DeclareMathOperator{\E}{\mathrm{E}}
\DeclareMathOperator{\R}{\mathrm{R}}
\DeclareMathOperator{\Rh}{\mathrm{\hat R}}
\DeclareMathOperator{\argmin}{\mathrm{arg\text{\,}min}}
\DeclareMathOperator{\tr}{tr}

\DeclareMathOperator{\I}{\mathrm{I}}
\DeclareMathOperator{\Dy}{\mathrm{D_y}}

\DeclareMathOperator{\F}{\mathrm{F}}
\DeclareMathOperator{\Fh}{\mathrm{\hat F}}
\DeclareMathOperator{\p}{\mathrm{p}}
\DeclareMathOperator{\Eh}{\mathrm{\hat E}}
\DeclareMathOperator{\q}{\mathrm{q}}

\DeclareMathOperator{\D}{\mathrm{D}}
\DeclareMathOperator{\Dw}{\mathrm{D_w}}

\newtheorem{theorem}{Theorem}
\newtheorem{example}{Example}
\newcommand{\CC}{\mathbb{C}}
\newcommand{\RR}{\mathbb{R}}
\newcommand{\arr}[2]{\begin{array}{#1} #2\end{array}}
\newcommand{\mat}[2]{\left[\!\!\arr{#1}{#2}\!\!\right]}

\begin{document}
\maketitle

\begin{abstract}
    We present an approach to backpropagating through minimal problem solvers in end-to-end neural network training. Traditional methods relying on manually constructed formulas, finite differences, and autograd are laborious, approximate, and unstable for complex minimal problem solvers. 
    We show that using the Implicit function theorem (IFT) to calculate derivatives to backpropagate through the solution of a minimal problem solver is simple, fast, and stable. We compare our approach to (i) using the standard autograd on minimal problem solvers and relate it to existing backpropagation formulas through SVD-based and Eig-based solvers and (ii) implementing the backprop with an existing PyTorch Deep Declarative Networks (DDN) framework~\cite{Gould2022}. We demonstrate our technique on a toy example of training outlier-rejection weights for 3D point registration and on a real application of training an outlier-rejection and RANSAC sampling network in image matching. Our method provides $100\%$ stability and is 10 times faster compared to autograd, which is unstable and slow, and compared to DDN, which is stable but also slow.  
\end{abstract}
\section{Introduction}\label{sec:Introduction}
Recently, minimal problem solvers~\cite{Nistr2004AnES,DBLP:conf/cvpr/SteweniusNKS05,kukelova2008automatic,Larsson-CVPR-2018,DBLP:conf/cvpr/MartyushevVP22} have been incorporated into end-to-end machine learning pipelines in camera localization~\cite{Brachmann2017}, image matching~\cite{Wei_2023_ICCV}, and geometric model estimation by RANSAC~\cite{Brachmann:2019}. The key problem with using minimal problem solvers in end-to-end neural network training is to make them differentiable for backpropagation. 

Early attempts to backpropagate through minimal problem solvers used explicit derivative formulas~\cite{Ionescu2015,DBLP:conf/eccv/DangYHWFS18,Ranftl2018} and finite differences~\cite{Brachmann2017} to compute derivatives for backpropagation. Recent work~\cite{Wei_2023_ICCV} proposed computing the derivatives using autograd. These are valid approaches, but they can still be considerably improved. Manual differentiation is laborious and must be done repeatedly for every new problem. Finite differences are approximate and are prone to numerical errors. Using autograd is also limited to relatively simple minimal problem solvers since, for more complex solvers with large templates~\cite{DBLP:conf/cvpr/MartyushevVP22}, differentiating the templates becomes unstable due to, e.g., vanishing of the gradients~\cite{DBLP:journals/tnn/BengioSF94}. 

\begin{figure}[t]
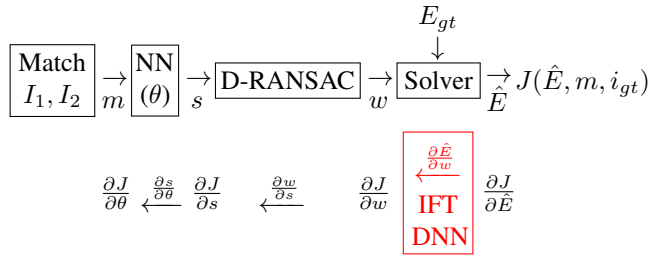

{\parbox{0.92\linewidth}{
\fbox{\parbox{0.9cm}{\centering Match \\ $I_1, I_2$}}
\parbox{0.3cm}{\centering $~$\\$\rightarrow$\\[-0.3em]$m$}
\fbox{\parbox{0.4cm}{\centering \!NN\\($\theta$)}}
\parbox{0.3cm}{\centering $~$\\$\rightarrow$\\[-0.3em]$s$}
\fbox{\parbox{1.7cm}{\centering \!D-RANSAC\!}}
\parbox{0.3cm}{\centering $~$\\$\rightarrow$\\[-0.3em]$w$}
\parbox[b]{1.1cm}{\centering $E_{gt}$\\$\downarrow$\\\fbox{\parbox{0.88cm}{\centering \!Solver\!}}}
\parbox{0.3cm}{\centering $~$\\$\rightarrow$\\[-0.1em]$\hat{E}$}
\parbox{0.9cm}{$J(\hat{E},m,i_{gt})$}\\[1ex]
\hspace*{1.1cm}
{\parbox{0.45cm}{\centering $\frac{\partial J}{\partial \theta}$}}
{\parbox{0.6cm}{\centering $\xleftarrow{\frac{\partial s}{\partial \theta}}$}}
{\parbox{0.4cm}{\centering $\frac{\partial J}{\partial s}$}}
{\parbox{1.45cm}{\centering $\xleftarrow{\frac{\partial w}{\partial s}}$}}
{\parbox{0.8cm}{\centering $\frac{\partial J}{\partial w}$}}
\hspace*{-0.1cm}{\color{red} \fbox{\parbox{0.7cm}{\centering $\xleftarrow{\frac{\partial \hat{E}}{\partial w}}$ IFT\\ DNN}}}
\hspace*{-0.2cm}{\parbox{0.9cm}{\centering $\frac{\partial J}{\partial \hat{E}}$}}
}}
\caption{A typical end-to-end training pipeline with a minimal problem Solver~\cite{Wei_2023_ICCV} that trains a Neural Network (NN) to predict correct matches. Forward pass: Tentative handcrafted matches $m$ between images $I_1, I_2$ are clarified using NN~\cite{zhao2021progressive} parametrized by trained weights $\theta$, and scores $s$ for these matches are computed. Differentiable D-RANSAC selects a minimal data sample $w$ using the scores $s$; the Solver computes a model $\hat{E}$, which is scored by the loss $J$ using the correct matches $m(i_{gt})$ with the ground truth inlier indicator $i_{gt}$. The groundtruth $E_{gt}$ is passed to the Solver to choose the closest model. Backward pass: Gradient $\frac{\partial J}{\partial \theta}$ for training weights $\theta$ is computed by the chain rule. {\em The key issue is robustly and efficiently backpropagating through the Solver. We propose to use the Implicit function theorem (IFT) directly or implement the backpropagation via the PyTorch Deep Declarative Network (DDN) machinery~\cite{Gould2022}.}}
\label{fig:DFRpipeline}
\end{figure}

Let us illustrate using the minimal problem solvers in end-to-end neural network training for calibrated image matching~\cite{Wei_2023_ICCV}, Fig.~\ref{fig:DFRpipeline}. First, a for pair of~images $I_1, I_2$ the RootSIFT~\cite{6248018} pre-detected features are used to produce tentative matches $m$, and the~Neural Network $s=\mbox{NN}(\theta;m)$ computes scores~$s$ of the~matches~$m$. The network is parameterized by weights $\theta$, the learned pipeline parameters. Next, the~differentiable $w = \mbox{D-RANSAC}(m,s)$ selects a~minimal data sample~$w$ from $m$ using the $s$ scores. The sample $w$ is passed to the Essential matrix solver $E = \mbox{Solver}(w)$, which computes up to 10 $E$ candidates and selects the best $\hat{E}$ that is closer to the known ground truth $E_{gt}$. The Solver can thus be seen as solving a discrete optimization problem consisting of first solving a set of polynomial equations and then choosing the optimal one w.r.t.\ $E_{gt}$. Having $\hat{E}$, a projection-based loss $J(\hat{E}, M, i_{gt})$ is calculated from all tentative matches $m$ and the indicator $i_{gt}$ of the known correct (inlier) matches. As a function to be optimized, the pipeline is the composition 
\begin{equation*}
    J(w;E_{gt},i_{gt}) = J(\hat{E}; m,i_{gt}) \circ \hat{E}(w;E_{gt}) \circ w(s;m) \circ s(\theta;m)
\end{equation*}
where $E_{gt}$, $i_{gt}$, and $m$ can be seen as parameters provided by training data. 

To learn weights $\theta$, the (transposed) gradient 
\begin{equation*}
    \frac{\partial L}{\partial \theta} = 
    \frac{\partial L}{\partial \hat{E}}
    \frac{\partial \hat{E}}{\partial w}
    \frac{\partial w}{\partial s}
    \frac{\partial s}{\partial \theta}
\end{equation*}
of $J$ w.r.t.\ $\theta$ has to be calculated. Calculating
$\frac{\partial L}{\partial \hat{E}}, \frac{\partial w}{\partial s}, \frac{\partial s}{\partial \theta}$ is straightforward and efficient using autograd. Calculating $\frac{\partial \hat{E}}{\partial w}$ could also be attempted by autograd. We show that it often fails even for a simple minimal problem of computing the calibrated relative camera pose from five image matches~\cite{Nistr2004AnES}. Our main contribution is to show how to calculate $\frac{\partial \hat{E}}{\partial w}$ robustly by the existing PyTorch machinery of Deep Declarative Networks (DDN)~\cite{Gould2022} and {\em robustly an efficienlty} using the Implicit function theorem (IFT).

\subsection{Motivation}\label{sec:Motivation}

Minimal problem solvers can be complex. Symbolic-numeric minimal problem solvers~\cite{Elkadi2005,Nistr2004AnES,Larsson-CVPR-2018, DBLP:conf/cvpr/MartyushevVP22} consist of potentially very large templates (formulas combined with the Gauss-Jordan elimination) to construct a matrix whose eigenvectors provide the solutions. The size of the templates is, for systems with a finite number of solutions and more than three unknowns,  proportional to a polynomial in $D^n$, where $D$ is the mean degree of input constraints and $n$ is the number of unknowns~\cite{doi:10.1142/S0218196711006364}. In practice, it is not always necessary to construct a full Groebner basis~\cite{Cox-IVA-2015}, but practical automatic solver generation algorithms~\cite{kukelova2008automatic, larsson2017making,DBLP:conf/cvpr/MartyushevVP22} still produce very large templates for problems with more variables. See, for example, the sizes of the templates in~\cite{DBLP:conf/cvpr/MartyushevVP22} and the related discussion there. 

Therefore, backpropagating through such templates using autograd on the fully expanded computational graph, as in~\cite{Wei_2023_ICCV}, is generally slower and potentially less stable. Another issue is the backpropagation stability through large eigenvector decompositions~\cite{DBLP:conf/eccv/DangYHWFS18}. Backpropagation through constraints at the optimum using the Implicit function theorem (for active KKT constraints in semi-algebraic cases) is much simpler, with linear complexity in the size of the input constraints, and is stable.

\subsection{Contribution}\label{sec:Contribution}

We investigate backpropagation methods through minimal problem solvers in end-to-end neural network training. We show that using the Implicit function theorem to calculate derivatives to backpropagate through the~solution of a minimal problem solver is simple, fast, and stable. We provide a direct implementation of IFT backpropagation that is stable and fast. We also show how a stable but slower backpropagation can be implemented using the existing PyTorch Deep Declarative Networks framework~\cite{Gould2022}. This second approach may be helpful for quick testing of functionality before a more efficient IFT backpropagation is implemented. 

We compare our approach to using the standard autograd on minimal problem solvers~\cite{Wei_2023_ICCV} and relate it to the existing formulas for backpropagating through SVD-based and Eig-based solvers~\cite{Ionescu2015, DBLP:conf/eccv/DangYHWFS18}. 

We demonstrate our technique on toy examples of training outlier-rejection weights for 3D point registration and on a real application of training an outlier-rejection and RANSAC sampling network~\cite{Wei_2023_ICCV} in two image matching and camera relative pose computation. Our IFT backpropagation provides $100\%$ stability and is 10 times faster than unstable autograd and DDN.

\section{Previous work}\label{sec:PreviousWork}

The interest in backpropagating through minimal problem solvers started with attempting to develop a differentiable version of RANSAC that could be used in end-to-end learning pipelines. 

In seminal work~\cite{Brachmann2017}, a differentiable pipeline for camera localization, including RANSAC and PNP solver, has been suggested. The PNP solver was so efficient that it was possible to estimate the derivatives of its output by numerical central differences. We show in Example~\ref{ex:P3P} how to easily estimate the derivatives using the Implicit function theorem.  

In~\cite{Ranftl2018}, a method for robust estimation of fundamental matrices embedded in an end-to-end training pipeline was proposed. The goal was to gradually learn weights for scoring tentative matches to suppress mismatches in a sequence of weighted least-squares problems. Here, the fundamental matrices are computed via SVD. Backpropagating through SVD uses derivatives computed by explicit formulas derived  in~\cite{Ionescu2015}. Computing derivatives explicitly by formulas is the best approach, but it can be laborious when done manually, as in~\cite{Ionescu2015}. We use this example to show in Section~\ref{sec:Exp:3D} that our approach based on DDN/IFT provides equally correct results.

Recent work~\cite{Wei_2023_ICCV} presents another differentiable pipeline for image matching that includes a differentiable version of RANSAC and minimal problem solvers for fundamental and essential matrix computation. Here, the solvers may use minimal data samples, i.e., \ 7 and 5 correspondences, respectively, and solve for the matrices using symbolic-numeric algebraic minimal problem solvers, e.g.~\cite{Nistr2004AnES}. To backpropagate through the minimal problem solvers,~\cite{Wei_2023_ICCV} uses autograd on the templates and the following Eigendecomposition. As explained above, this is a possible approach for small templates but should be replaced by our approach whenever possible. We show in experiments, Section~\ref{sec:Exp:EG}, that using autograd often fails even for the relatively simple minimal problem solver of essential matrix estimation. In contrast, our approach based on using the Implicit function theorem delivers $100\%$ stable results and is 10 times faster. 

Our approach relies on using the Implicit function theorem~\cite{Rudin-1976}, which was already used to implement backpropagation for SVD and Eigendecomposition~\cite{Ionescu2015,DBLP:conf/eccv/DangYHWFS18} computation blocks of learning pipelines. These approaches manually compute formulas for Jacobians and gradients and implement backward passes for the modules. It is the~simplest, fastest, and most robust approach currently possible. It~should be used in all final implementations. 

To provide a simple but slow error-prone engineering solution for implementing backpropagation through algebraic solvers, we exploit the DDN~\cite{Gould2022} that implement a framework to backpropagate through algebraic and semi-algebraic optimizers. A similar approach has been suggested in~\cite{DBLP:journals/corr/abs-2003-01822}, but we prefer~\cite{Gould2022} since it provides PyTorch framework implementation.

\subsection{IFT end-to-end learning}

The implicit function theorem (IFT) was used in several interesting works related to using geometric optimization in end-to-end learning. None of the works deals with minimal problems, but they are very related to~the~tasks that are solved with minimal problem solvers, and thus we comment on them here. 

In~\cite{DBLP:conf/cvpr/0009PCLC20}, IFT is used to backpropagate through the least-squares PNP optimization solver for end-to-end learnable geometric vision. Explicit formulas for the Jacobians of~the~PNP least-squares solvers are derived, and it is shown that they lead to accurate and stable backpropagation. This is an interesting work because it presents a particular generalization of polynomial problems to rational problems but concentrates on one specific geometrical problem.  

In~\cite{DBLP:conf/cvpr/ChenWWTX022}, there is a method for backpropagation through a~probabilistic PNP. Technically, this approach does not use IFT since the PNP in this formulation is not a solution to~equations, but a function providing a distribution over the~domain of~camera poses. The~derivatives of the output are thus readily computable by~autograd or by~explicit formulas. 

In~\cite{DBLP:conf/icml/AmosK17}, the OptNet layer for propagating through optimization problems solved by layers of small quadratic programs is developed using IFT. The aim is to provide an efficient and stable backpropagation that can be implemented on GPU. This is an interesting work since it shows how to backpropagate through an~important class of optimization problems. However, it can't address general minimal problems since it only addresses the problems specified as a quadratic program with linear constraints.  

\section{Using the Implicit function theorem}\label{sec:BPbyIFT}
Let us now explain and demonstrate how to compute derivatives of the output of a minimal problem solver using the Implicit function theorem.

Technically, minimal problem solvers generate solutions ${\bf x}({\bf a})$ to systems of polynomial equations $f_1({\bf a}),\ldots,f_K({\bf a})$ based on input parameters ${\bf a}$. To backpropagate through the minimal problem solvers, it is necessary and sufficient to compute the derivatives $\frac{\partial {\bf x}}{\partial {\bf a}}$ at each solution ${\bf x}$ w.r.t.\ parameters ${\bf a}$. 

We now explain how the Implicit function theorem~\cite{Rudin-1976} can be used to implement simple, efficient, and numerically stable computation of the derivatives of ${\bf x}$ w.r.t.\ ${\bf a}$. Let us start with the formulation of the theorem itself, and then consider a simple example. \\
\newsavebox{\thmdxda}
\sbox{\thmdxda}{%
\parbox{\linewidth}{%
\begin{theorem}[Differentiating roots of a polynomial system]
Let $H({\bf a}) = h_1({\bf x},{\bf a}),\ldots,h_K({\bf x},{\bf a})$ be a sequence of $K$ complex polynomials in $N$ unknowns ${\bf x} = x_1,\ldots,x_N$ and $M$ unknowns ${\bf a} = a_1,\ldots,a_M$. Let ${\bf x}({\bf b})$ be an isolated multiplicity-one solution to $H({\bf a})$ in ${\bf x}$ for ${\bf a} := {\bf b}\in \CC^M$. Then, 
{\small
\[
\left[\frac{\partial x_n({\bf a})}{\partial a_m}({\bf b})\right] \!\!=\!
- \left[\frac{\partial h_k({\bf x},{\bf a})}{\partial x_n}({\bf x}({\bf b}),{\bf b})\right]^{\!+}\!\!
\left[\frac{\partial h_k({\bf x},{\bf a})}{\partial a_m}({\bf x}({\bf b}),{\bf b})\right] 
\]
}
with $k=1,\ldots,K$, $m=1,\ldots,M$, $n=1,\ldots,N$, and $[A]^+$ denoting the pseudoinverse~\cite{Meyer-2001} of the (Jacobian) matrix $[A]$. 
\label{thm:dx/da}
\end{theorem}
}}
\usebox{\thmdxda}
\vspace*{-2em}
\begin{proof}
Informal statement: The Implicit function theorem is a very standard fact. Here we use it in a special situation when functions are polynomials, since we focus on the Computer Vision problems. The main issue is thus ensuring that the assumptions of the Implicit function theorem are satisfied. For polynomials, which are continuous, infinitely differentiable functions, the assumptions of the Implicit function theorem are satisfied iff the solutions to the polynomials are isolated and have multiplicity one. See Supplementary Material, Section~\ref{sec:dx/da} for more details. 
\end{proof}
Let us demonstrate the above theorem with an example of the classical computer vision P3P minimal problem of solving the absolute pose of the calibrated camera~\cite{DBLP:journals/cacm/FischlerB81}. 
\begin{example}\label{ex:P3P}
Consider a general configuration of three 3D points ${\bf A}_1, {\bf A}_2, {\bf A}_3 \in \RR^3$, their respective calibrated image projections represented by homogeneous coordinates ${\bf a}_1, {\bf a}_2, {\bf a}_3 \in \RR^3$ and the vector  ${\bf x} = [x_1, x_2,x_3]^\top \in \CC^3$ of the depths of the points.
Stack all 3D points and image points in a single vector ${\bf a} = [{\bf A}_1; {\bf A}_2; {\bf A}_3; {\bf a}_1; {\bf a}_2; {\bf a}_3] \in \RR^{18}$. The 3D points, their projections, and depths are related by 
\begin{eqnarray*}
0 = h_1({\bf x},{\bf a}) = ||{\bf A}_1-{\bf A}_2||^2 - ||x_1 {\bf a_1}-x_2 {\bf a}_2||^2 \\
0 = h_2({\bf x},{\bf a}) = ||{\bf A}_2-{\bf A}_3||^2 - ||x_2 {\bf a_2}-x_3 {\bf a}_3||^2 \\
0 = h_3({\bf x},{\bf a}) = ||{\bf A}_3-{\bf A}_1||^2 - ||x_3 {\bf a_3}-x_1 {\bf a}_1||^2
\end{eqnarray*}
The calibrated absolute camera pose computation solves for the depths ${\bf x}({\bf a})$ as functions of the parameters ${\bf a}$. In this example, $K=3$, $N=3$, and $M = 18$. The polynomial system above generically has $8$ multiplicity-one solutions ${\bf x}_s({\bf a}), s=1,\ldots,8$~\cite{DBLP:journals/cacm/FischlerB81}. Now, we compute the derivatives of the solutions for a generic parameter vector ${\bf b}$. Let us first compute the derivatives of the functions $h_k({\bf x},{\bf a})$ w.r.t\ ${\bf x}$ and~${\bf a}$
{\tiny
\begin{multline*}
J_{\bf x}({\bf x},{\bf a}) = \left[\frac{\partial h_k({\bf x},{\bf a})}{\partial x_n}\right] = 
\mat{ccc}{\frac{\partial h_1}{\partial x_1}&\frac{\partial h_1}{\partial x_2}&\frac{\partial h_1}{\partial x_3} \\[2ex]
          \frac{\partial h_2}{\partial x_1}&\frac{\partial h_2}{\partial x_2}&\frac{\partial h_2}{\partial x_3} \\[2ex]
          \frac{\partial h_3}{\partial x_1}&\frac{\partial h_3}{\partial x_2}&\frac{\partial h_3}{\partial x_3}}({\bf x},{\bf a}) = \\
= 2\mat{ccc}{-{\bf a}_1^\top(x_1 {\bf a_1}-x_2 {\bf a}_2)&\phantom{-}{\bf a}_2^\top(x_1 {\bf a_1}-x_2 {\bf a}_2)&0\\\
             0&-{\bf a}_2^\top(x_2 {\bf a_2}-x_3 {\bf a}_3)&\phantom{-}{\bf a}_3^\top(x_2 {\bf a_2}-x_3 {\bf a}_3)\\
             \phantom{-}{\bf a}_1^\top(x_3 {\bf a_3}-x_1 {\bf a}_1)&0&-{\bf a}_3^\top(x_3 {\bf a_3}-x_1 {\bf a}_1)} \\
J_{\bf a}({\bf x},{\bf a}) = \left[\frac{\partial h_k({\bf x},{\bf a})}{\partial a_m}\right] = 
\mat{ccc}{\frac{\partial h_1}{\partial a_1}&\ldots&\frac{\partial h_1}{\partial a_{18}} \\
\vdots & \ddots & \vdots \\
\frac{\partial h_3}{\partial a_1}&\ldots& \frac{\partial h_3}{\partial a_{18}}}({\bf x},{\bf a}) = \left[K_1 | K_2\right]\\
K_1 = 2\mat{ccc}{{\bf A}_1^\top-{\bf A}_2^\top&{\bf A}_2^\top-{\bf A}_1^\top&0\\
0&{\bf A}_2^\top-{\bf A}_3^\top&{\bf A}_3^\top-{\bf A}_2^\top\\
{\bf A}_1^\top-{\bf A}_3^\top& 0 & {\bf A}_3^\top-{\bf A}_1^\top}\\
K_2 = 2\mat{ccc}{-x_1(x_1 {\bf a_1}-x_2 {\bf a}_2)&\phantom{-}x_2(x_1 {\bf a_1}-x_2 {\bf a}_2)&0\\
0&-x_2(x_2 {\bf a_2}-x_3 {\bf a}_3)&\phantom{-}x_3(x_2 {\bf a_2}-x_3 {\bf a}_3)\\
\phantom{-}x_1(x_3 {\bf a_3}-x_1 {\bf a}_1)&0&-x_3(x_3 {\bf a_3}-x_1 {\bf a}_1)}
\end{multline*}}
Next we evaluate $J_{\bf x}({\bf x},{\bf a})$ and $J_{\bf a}({\bf x},{\bf a})$ at a generic parameter vector ${\bf b} = [0,0,3,  2,0,3, 0,6,3,-1/3,-1/3,1, \newline 1/3,-1/3,1,  -1/3,5/3,1] $ and, e.g., the first solution ${\bf \hat{x}}({\bf b}) = [3,3,3]$
{\tiny
\begin{eqnarray*}
J_{\bf x}({\bf \hat{x}}({\bf b}),{\bf b}) &=& \mat{rrr}{-1.33&-1.33&0.00\\0.00&-5.33&-21.33\\-4.00&0.00&-20.00} \\
J_{\bf a}({\bf \hat{x}}({\bf b}),{\bf b}) &=& 
\mat{rrrrrrrrrr}{-4&0  &0 & 4&0  &0 &\ldots&   0&  0&0\\
                  0&0  &0 & 4&-12&0 &\ldots&  12&-36&0\\
                  0&-12&0 & 0&  0&0 &\ldots&   0&-36&0\\}
\end{eqnarray*}
}
and compute 
{\tiny
\begin{eqnarray*}
    \left[\frac{\partial \hat{x}_n({\bf a})}{\partial a_m}({\bf b})\right] &=& 
    -J_{\bf x}({\bf \hat{x}}({\bf b}),{\bf b})^{-1} 
    J_{\bf a}({\bf \hat{x}}({\bf b}),{\bf b}) = \\ 
    &=&
    \mat{rrrrrrr}{1.66& 1.33 &0&\ldots& 1.25& 0.24&0\\
                  1.33&-1.33 &0&\ldots&-1.25&-0.24&0\\
                 -0.33& 0.33 &0&\ldots&-0.25&-1.75&0} \text{ .}
\end{eqnarray*}}
\end{example}

\section{Implementing MinBackProp with IFT}\label{sec:MBPbyIFT}

The first option to implement the backpropagation for a~minimal problem solver is to use the Implicit function theorem in a straightforward way (as in Example~\ref{ex:P3P}). This has been done before, e.\,g., in~\cite{Ionescu2015} for SVD and EIG solvers. To use the Implicit function theorem directly, one can construct a system of polynomial equations, compute the derivatives of the system w.\,r.\,t. inputs and outputs manually or with the help of a computer algebra system~\cite{Maple}, and follow the~\cref{thm:dx/da} to compute the Jacobian of the output with respect to the input. In~case the system of equations is overdetermined, one can write down the Lagrange multipliers method and then use the IFT for the new system. 

Applying the~Implicit function theorem involves inverting the Jacobian of the polynomial system w.\,r.\,t. the~output, so before inverting the Jacobian one can check if the matrix is full-rank using SVD. If~it~is~not, one can randomly choose a full-rank Jacobian from~the~batch to~invert.

\section{Implementing MinBackProp with DDN}\label{sec:MBPbyDDN}
The second option to backpropagate through a minimal problem solver is to use a fully automatic method to~compute the~derivatives.
The Deep Declarative Networks (DDN) framework, introduced in~\cite{Gould2022}, provides a PyTorch implementation allowing backpropagation through optimization solvers, including algebraic, semi-algebraic (i.e., \ with inequalities), and non-polynomial problems. Next, we explain how to use the~DDN framework for implementing backpropagation for~minimal problem solvers. 
\\[1ex]\noindent{\bf Declarative node:} 
The DDN framework introduces declarative nodes that take input $\mathrm{w}$ into the optimal solutions
\[
{\hat{y}}(\mathrm{w}) = \underset{\mathrm{y} \in \mathrm{C}}{\argmin}\, f(\mathrm{w}, \mathrm{y})
\]
where $C$ is a constraint set.
\\[1ex]\noindent{\bf Constraint set $C$:} For algebraic minimal problem solvers, the set $C = \{y \colon h_1(y)=0,\ldots,h_K(y)=0\}$ is an algebraic variety~\cite{Cox-IVA-2015} consisting of a finite number of points. Hence, in this case, the algebraic variety is a finite set of solutions to some system of polynomial equations $H = [h_1,\ldots,h_K]$. 
Assuming the genericity of data, which is often guaranteed by having noisy measurements, solutions to $H$ have all multiplicity equal to one. Hence it is easy to compute derivatives of the solution to $H$ w.r.t.\ parameters as demonstrated in Section~\ref{sec:BPbyIFT}. 
\\[1ex]\noindent{\bf Loss $f$:}  The low-level loss depends on a particular problem. DDN framework implements backpropagation for a general declarative node and thus expects a~loss function $f(\mathrm{w},\mathrm{y})$ with non-degenerate derivatives. For instance, in Section~\ref{sec:Exp:EG}, thus, one must ``invent'' such a loss function for minimal problem solvers since they provide exact solutions that always satisfy equations $H$ exactly. A natural choice is to use the sum of squares of the equation residuals from a subset of $H$, i.e., \ $f = \sum_{h_i \in G \subseteq H} h_i^2$, such that the derivatives of $f$ are~non-degenerate. A possible choice is to use a~random linear combination of some $h_i$'s. For instance, in Section~\ref{sec:Exp:EG}, we use the sum of squares of the linear constraints only to keep $f$ simple and non-degenerate. 

It is important to mention that in the pipeline of~\cite{Wei_2023_ICCV}, the ``Solver'' not only computes the~solutions $\mathrm{y}$ to a polynomial system $H$ but also selects the best solution $\mathrm{\hat{y}}$ by finding  $\mathrm{\hat{y}}$ that is closest to~the~ground-truth $\mathrm{y}_{gt}$ in $\|\mathrm{\hat{y}}-\mathrm{y}_{gt}\|^2$ sense. This discrete optimization step does not influence the derivatives of $\mathrm{\hat{y}}$ w.r.t\ $\mathrm{w}$ since, locally, small changes of $\mathrm{w}$ do not influence the~result of this discrete optimization as small changes to $\mathrm{w}$ generically do not change the result of the discrete optimization.  
\\[1ex]\noindent{\bf End-to-end training formulation:}
End-to-end training of a pipeline with a declarative node is in~\cite{Gould2022} formulated as the following bi-level optimization problem
\begin{equation}
\begin{aligned}
\underset{\mathrm{w}}{\min}\,& J(\mathrm{w}, \mathrm{\hat y}(\mathrm{w})) \\
\text{s.\,t. } &\mathrm{\hat y} (\mathrm{w}) \in \underset{\mathrm{y \text{\,} \in \text{\,} \mathrm{C}}}{\argmin}\,
    f(\mathrm{w}, \mathrm{y}) \\
    & \hspace{1.5cm} \text{with } 
    C = \{\mathrm{y} \colon h(\mathrm{y})=0\}\text{ ,}
\end{aligned}
\label{eq:formulation}
\end{equation}
where $J(\mathrm{w}, \mathrm{y})$ is an upper-level loss, $f(\mathrm{w}, \mathrm{y})$ is a low-level loss, $h(\mathrm{y})$ are constraints, $\mathrm{w}$ is the input parameters to the declarative node, $\mathrm{y}$ is the model to~be~estimated, and $\mathrm{C}$ is a constraint set. The relation $\mathrm{w} \mapsto \mathrm{\hat y}$ is~defined as a solution to the low-level optimization problem provided by the declarative node. 
\\[1ex]\noindent{\bf Gradients for minimal problem solvers:}
To minimize the loss $J(\mathrm{w}, \mathrm{\hat{y}})$ in~(\ref{eq:formulation}) via the gradient descent, we need to compute $\D J(\mathrm{w}, \mathrm{\hat{y}})$\footnote%
{
Here we use the notation from~\cite{Gould2022} where $\D J(\mathrm{w}, \mathrm{\hat{y}})$ denotes the total derivative $\odv{J(\mathrm{w}, \mathrm{\hat{y}}(\mathrm{w}))}{\mathrm{w}}$ of loss $J(\mathrm{w}, \mathrm{\hat{y}}(\mathrm{w}))$ w.r.t.\ $\mathrm{w}$ when $\mathrm{\hat{y}}(\mathrm{w})$ is considered a function of $\mathrm{w}$, $\Dw J(\mathrm{w}, \mathrm{\hat{y}})$ means partial derivatives $\frac{\partial J(\mathrm{w}, \mathrm{\hat{y}})}{\partial \mathrm{w}}$ of $J(\mathrm{w}, \mathrm{\hat{y}})$ w.r.t.\ $\mathrm{w}$ when $\mathrm{\hat{y}}$ is consider fixed, and $\D_{\mathrm{\hat{y}}} J(\mathrm{w}, \mathrm{\hat{y}})$ means partial derivatives $\frac{\partial J(\mathrm{w}, \mathrm{\hat{y}})}{\partial \mathrm{\hat{y}}}$ of $J(\mathrm{w}, \mathrm{\hat{y}})$ w.r.t.\ $\mathrm{\hat{y}}$ when $\mathrm{w}$ is consider fixed.
}. 
 In our formulation, the~loss function $J(\mathrm{w}, \mathrm{\hat{y}})$ does not depend on $\mathrm{w}$ explicitly. Therefore
\begin{align}
    \D J(\mathrm{w}, \mathrm{\hat{y}}) &=  \Dy J(\mathrm{w}, \mathrm{\hat{y}}) \D \mathrm{\hat{y}}(\mathrm{w})   \text{ .}
\end{align}
The gradients $\Dy J(\mathrm{w}, \mathrm{\hat{y}})$ are the gradients computed before the declarative block and the $\D \mathrm{\hat{y}}(\mathrm{w})$ are the gradients through the solution of the low-level optimization problem. 
\\[1ex]\noindent{\bf Declarative node specification:}
The important feature of the DDN framework is that one does not have to provide formulas for the derivatives explicitly. It is enough to specify a declarative node by defining the~loss $f(\mathrm{w}, \mathrm{y})$ and the~constraints 
$H = [h_1,\ldots,h_K]$. The backward pass based on the derivatives is then automatically computed by the DDN framework. This makes the implementation of backpropagation through minimal problem solvers easy and robust.

\begin{figure}
    \centering
     \includegraphics[width=\linewidth]{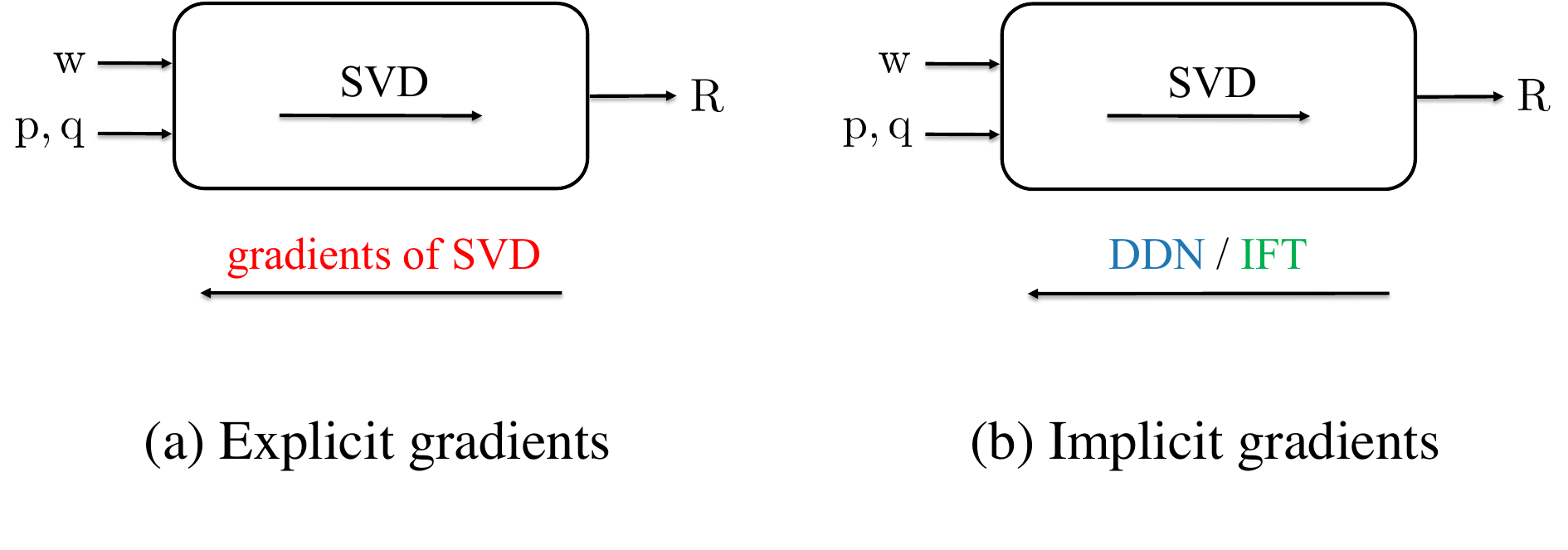}
    \caption{The figure demonstrates the backpropagation for~the~toy example for~the~3D point registration with an~outlier. The forward pass is~the~Kabsch algorithm~\cite{Kabsch:a12999} (SVD) and remains the~same for~both explicit and~implicit methods. The~backward pass is~performed explicitly via~the~closed-form gradients of SVD~\cite{Ionescu2015}~(a) and~implicitly, using the Deep Declarative Networks (DDN)~\cite{Gould2022}~and the Implicit function theorem~(IFT)~(b).  }
    \label{fig:ddn_svd_backprop_ab}
\end{figure}

\begin{figure*}
    \centering
    \includegraphics[width=1.01\textwidth]{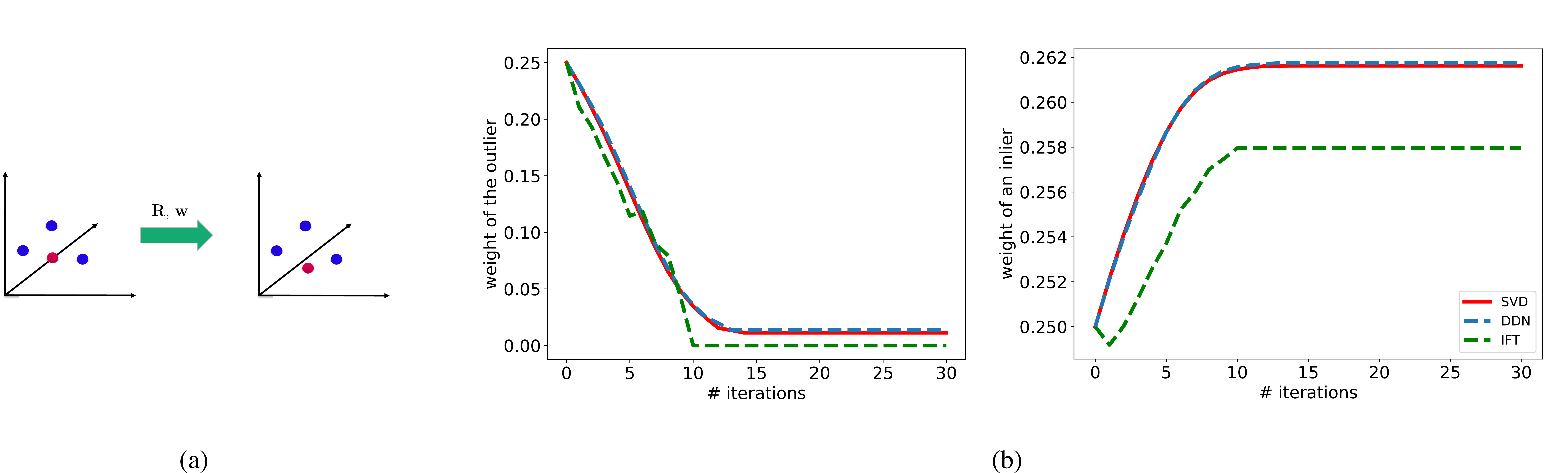}
    \caption{A toy example for the 3D point registration problem with an outlier: (a) visualization of~four points before and~after rotation~$\R$ with~inlier/outlier weights $\mathrm{w}$, blue points denote inliers, and the magenta point is the~outlier; (b)~the~weight for~the~outlier~(left) and the~weight for an~inlier~(right) during the~optimization. The~red color indicates explicit computation of~the~gradients, the~blue color denotes backpropagation with~the~DDN, and the~green color is backpropagation with the~IFT. One can see that the~gradients computed with~the~DDN and the~IFT approximate the~gradients of~SVD computed in~closed~form.}
    \label{fig:3dreg-w0-w1}
\end{figure*}

\section{Experiments}\label{sec:Expatiments}
In our experiments, we compute the $\D \mathrm{y}(\mathrm{w})$ for a solution of a minimal problem in three ways. First, we compute the gradients using the closed-form solution and backpropagate explicitly using autograd. Secondly, we compute the~gradients $\D \mathrm{y}(\mathrm{w})$ via the Implicit function theorem and backpropagate through a solution by using them directly. Finally, we use the Deep Declarative Networks framework to implement the backpropagation by specifying a suitable optimization problem. Our experiments show that the derivatives of a solution to the minimal problem can be computed implicitly; they accurately approximate the derivatives computed explicitly, leading to more stable results. 

We consider two types of experiments to show the effectiveness of the proposed method. Firstly, we consider a~toy example for the 3D point registration with an outlier. We~chose this task since a solution to the~minimal problem during the forward pass can be computed using SVD, and the gradients of SVD can be computed both explicitly, using the closed-form solution, and implicitly, using the~Implicit function theorem and~the~Deep Declarative Networks. Secondly, we incorporate our MinBackProp into the existing state-of-the-art epipolar geometry framework, backpropagate through it, and compare the results on the real data. 

Our main goal is to show that we obtain equivalent behavior of the whole learning pipeline as the baseline~\cite{Wei_2023_ICCV} but in a more stable and efficient way. Note that we do not aim to improve the learning task of~\cite{Wei_2023_ICCV} but demonstrate better computational machinery for solving it.  





\subsection{3D Point Registration with an Outlier}\label{sec:Exp:3D}
Given two sets $P$ and $Q$ of $3$D points, a point $\mathrm{p_i} \in P$ corresponds to the point $\mathrm{q_i} \in Q$, $i = 1,\ldots,N$, where $N$ is the number of points. We want to find a rotation matrix $\R \in \mathbb{R}^{3 \times 3}$: $\mathrm{q_i} = \R \mathrm{p_i}$, $\forall i$. Given the ground-truth rotation matrix $\mathrm{R_{true}} \in \mathbb{R}^{3 \times 3}$, we define the weights $w_i$, $i = 1,\ldots,N$, for each correspondence $\mathrm{p_i} \leftrightarrow \mathrm{q_i}$ to~mark inliers and~outliers. 
To find $\R$ and $\mathrm{w} = [w_1, ..., w_N]$, let us consider a~bi-level optimization problem:
\begin{equation}
\begin{aligned}
\underset{\mathrm{w}}{\min}\,&J(\Rh (\mathrm{w})) \\
\text{s.\,t. } &\Rh (\mathrm{w}) \in \underset{\mathrm{R \text{\,} \in \text{\,} SO(3)}}{\argmin}\,
    f(\mathrm{w}, \R) \\
    & \hspace{1.3cm} \text{s.\,t. } h(\R) = 0 \text{ .}
\end{aligned}
\label{eq:3d_point_main}
\end{equation}
Where $J(\Rh (\mathrm{w}))$ is the upper-lever loss:
\begin{equation}
    J(\Rh (\mathrm{w})) = \arccos \left( \frac{\tr \left( \Rh (\mathrm{w})\text{\,} \mathrm{R_{true}^{T}} \right) - 1}{2} \right) \text{ ,}
    \label{eq:geometric}
\end{equation}
measuring the angle of the residual rotation~\cite{HZ-2003} and 
\begin{equation}
f(P, Q, \mathrm{w}, \R) = f(\mathrm{w}, \R)  = \frac{1}{N} \sum_{i=1}^{N} w_i \left\| \R \mathrm{p_i} - \mathrm{q_i} \right\|_2^2 \text{ ,}
\label{eq:3d-low-level}
\end{equation}
is the low-lever loss, and $h(\R)$ is as the constraint 
\begin{equation}
h(\R) = \R^T \R - \I = 0 \text{ .}
\label{eq:3d-constraints}
\end{equation}
The low-level minimization problem (\ref{eq:3d-low-level}) is known as Wahba's problem~\cite{Wahba1965}. \\
\textbf{Forward pass.} During the forward pass, we solve the~low-level optimization problem and compute the~value of $J(\Rh (\mathrm{w}))$. The low-level optimization problem is solved by~the~Kabsch algorithm \cite{Kabsch:a12999}, which includes the computation of SVD of a matrix constructed from coordinates $P$ and $Q$. The forward pass is the same for both DDN/IFT and SVD layers~(Fig.~\ref{fig:ddn_svd_backprop_ab}). \\
\textbf{Backward pass.} During the backward pass, we compute all the derivatives with respect to the input and parameters via the chain rule and update the vector $\mathrm{w}$. The backward pass is different for the~SVD and the~DDN/IFT layers, and we compare these three methods on the toy example below.  \\
\textbf{A toy example.} Let us consider a toy example to compare the gradients computed explicitly and implicitly given a forward pass fixed for both methods. Given four random points in 3D space, we~transformed the~points with an identity transform $\mathrm{R_{true}}$ and corrupted one correspondence, for instance $\mathrm{p_1} \leftrightarrow \mathrm{q_1}$. We~want to~find the~rotation matrix $\mathrm{R}$ along with inlier/outlier mask~$\mathrm{w}$ (see~Fig.~\ref{fig:3dreg-w0-w1}~(a)) given the correspondences $\mathrm{p_i} \leftrightarrow \mathrm{q_i}$. We~start with the~uniform distribution of~weights~$\mathrm{w}$ and~initialize them with~$1/N$. The developments of the values of $\mathrm{w}$ for an inlier and the outlier during the optimization are shown in~Fig.~\ref{fig:3dreg-w0-w1}~(b). We optimize the~weights $\mathrm{w}$ using the~gradient descent with the~learning rate of~$0.1$ for~$30$~iterations. The values of $\mathrm{w}$ for~the~outlier decrease to~$0$ and~the~values of~$\mathrm{w}$ for inliers increase. One can see that the gradients calculated implicitly approximate the gradients of SVD computed in closed form well.

One can find one more toy example for the fundamental matrix estimation (8-point) in the Supplementary Material,~\cref{supp:fundamental}.

\begin{table}
\centering
\resizebox{\columnwidth}{!}{%
\begin{tabular}{lccc}
\hline Method & AUC@$5^{\circ}$ & AUC@$10^{\circ}$ & AUC@$20^{\circ}$ \\ 
\hline
$\nabla$-RANSAC~\cite{Wei_2023_ICCV}  & 0.41 & 0.45 & 0.5 \\ 
MinBackProp DDN (ours) & 0.4 & 0.44 & 0.49  \\
MinBackProp IFT (ours) & 0.41 & 0.45 & 0.5  \\
\hline
\end{tabular}
}%
\caption{The average AUC scores for essential matrix estimation of $\nabla$-RANSAC~\cite{Wei_2023_ICCV} and MinBackProp (ours) over 12 scenes of the PhotoTourism dataset~\cite{PhotoTourism2020}, under different thresholds. Our method shows the same result as $\nabla$-RANSAC.}
\label{tab:essential}
\end{table}

\subsection{Training Outlier Detection in Epipolar Geometry Estimation}\label{sec:Exp:EG}
Our second experiment is incorporating our MinBackProp into the existing state-of-the-art epipolar geometry pipeline with real data. For that purpose, we consider a~fully-differentiable framework $\nabla$-RANSAC~\cite{Wei_2023_ICCV} that exploits the~5-point algorithm~\cite{Nistr2004AnES} for essential matrix estimation and evaluates it on the RootSIFT~\cite{6248018} features of the~PhotoTourism dataset~\cite{PhotoTourism2020}. Figure~\ref{fig:DFRpipeline} illustrates the architecture of~$\nabla$-RANSAC and the way how we integrate our MinBackProp in~it. We adopt all the formulations~and processing of~\cite{Wei_2023_ICCV}, except for the backpropagation through the minimal problem solver (red rectangle in Fig.~\ref{fig:DFRpipeline}). 

Given a pair of images $I_1$ and $I_2$ and a set of tentative correspondences $m = [\q_i, \tilde{\q}_i]$ computed from the RootSIFT features, the inlier indicator $i_{gt}$, and the ground truth essential matrix $\E_{gt}$. The goal is to train a Neural Network (NN)~\cite{zhao2021progressive} with parameters $\theta$ that predicts importance scores $s$ during inference. In Fig.~\ref{fig:DFRpipeline}, D-RANSAC indicates differentiable Gumbel Softmax sampler, $w$ denotes a minimal sample used as input into the Solver, and $J$ is the loss function. 

In accordance with~\cref{eq:formulation}, we formulate our bilevel optimization problem as follows. The upper-level objective $J$ is the symmetric epipolar distance over the inlier set $i_{gt}$ of matches $m_i = [\q_i, \tilde{\q}_i]$, $\Eh$ is the predicted model to be estimated, and $\tilde{l}_i = \Eh \q_i$, $l_i = \Eh^T \tilde{\q}_i$ are the epipolar lines
\begin{equation}
\begin{aligned}
J (\Eh, m, i_{gt}) &= \frac{1}{|i_{gt}|}\sum_{i \text{\,} \in \text{\,}i_{gt}} \left( \frac{1}{l_{i1}^2 + l_{i2}^2} + \frac{1}{\tilde{l}_{i1}^2 + \tilde{l}_{i2}^2} \right) \left( \tilde{\q}_i^T \Eh \q_i \right)^2 \text{,}
\end{aligned}
\label{eq:essential-upper-level}
\end{equation}
while the low-level objective is the ``invented'' loss consisting of the algebraic error of the constraints (which is always evaluated to zero but provides non-singular Jacobians, see the discussion in~\cref{sec:MBPbyDDN})
    \begin{equation}
    \begin{aligned}
    & \Eh (w) \in \underset{\text{\,} \E \in \text{\,} \mathbb{R}^{3 \times 3}}{\argmin}\,
        \frac{1}{5} \sum_{i=1}^{5} \left( \tilde{\q}_i^T \E \q_i \right)^2\\
        & \hspace{1.5cm} \text{s.\,t. } 2\E\E^T\E - \tr \left( \E\E^T \right)\E = 0 \text{ ,} \\
        & \hspace{4.67cm}   \left\| \E \right\|^2 = 1 \text{ ,}
    \end{aligned}
    \label{eq:essential-low-level}
    \end{equation}
where $w_i = [\q_i, \tilde{\q}_i]$ is a pair of matches from a minimal sample normalized by the intrinsic matrix pixel coordinates, and $\E$ is an essential matrix. The same equations~\cref{eq:essential-low-level} are used to construct the polynomial system for the Implicit function theorem and compute the~derivatives w.\,r.\,t. $\hat{E}$ and $[\q_i \text{, } \tilde{\q}_i]$ and the Jacobian $\frac{\partial \hat{E}}{\partial w}$, respectively. See Supplementary Material,~\cref{supp:essential} for more details.

As~in~the~\cref{sec:Exp:3D}, we~retain the~forward pass for the MinBackProp unchanged. However, we modify the~computation of gradients during the backward pass. Instead of computing the gradients explicitly by the autograd as in~$\nabla$-RANSAC, we exploit the Deep Declarative Networks and the Implicit function theorem. 

\begin{table}
\centering
\small{
\resizebox{\columnwidth}{!}{%
\begin{tabular}{lcc} 
\hline Method & stable runs ($\%$) & backward time (s) \\ 
\hline
$\nabla$-RANSAC~\cite{Wei_2023_ICCV} &  20 &  34.4 \\ 
$\nabla$-RANSAC$^{*}$ &  30 &  34.5 \\ 
MinBackProp DDN (ours) & \textbf{100} &  37.6 \\
MinBackProp IFT (ours) & \textbf{100} & \textbf{3.6} \\
\hline
\end{tabular}
}%
}
\caption{Stability metrics for 1K training of the baseline $\nabla$-RANSAC~\cite{Wei_2023_ICCV} and our MinBackProp methods for the essential matrix estimation. $\nabla$-RANSAC$^{*}$ denotes implementation of $\nabla$-RANSAC~\cite{Wei_2023_ICCV} without dropping non-real values in the solver during the forward pass. Our MinBackProp is much more stable compared to the $\nabla$-RANSAC and $\nabla$-RANSAC$^{*}$ and MinBackProp IFT ten times faster than the baseline.}
\label{tab:essential-stability}
\end{table}

For training, as in~$\nabla$-RANSAC, we use correspondences of~the~St.~Peter's Square scene of the~PhotoTourism dataset~\cite{PhotoTourism2020}, consisting of 4950 image pairs, with split $3:1$ into training and validation, while the rest 12 scenes remain for testing. 

To show the effect of our method, we perform two types of experiments. First, we want to show that our method does not decrease the baseline quality. Secondly, we want to show that computing the gradients via the Implicit function theorem is more stable and does not cause runtime errors during the backward pass when autograd tries to invert a singular matrix.

For the first experiment, the testing protocol is as follows: the scores $s$ are predicted by the trained NN, and the model $\E$ is estimated by MAGSAC++~\cite{barath2019magsacplusplus}. Table~\ref{tab:essential} shows the Area
Under the Recall curve (AUC), AUC~scores are averaged over 12 testing scenes for different thresholds. The essential matrix $\E$ is decomposed into $\R$ and $t$ using SVD and maximum error between $\R$ and $\R_{gt}$ and $t$ and $t_{gt}$ is reported. We train all three approaches for 10 epochs. One can see in Table~\ref{tab:essential} that our MinBackProp obtained the same scores as~the~$\nabla$-RANSAC, showing that it does not decrease the quality of the baseline. 

The second experiment aims to demonstrate the stability of the proposed MinBackProp compared to the~$\nabla$-RANSAC. For this purpose, we perform the~following experiment. Both methods are trained for 10 epochs, and 10 random runs are performed. We measure the~percentage of runs which finished 10 epochs training without any runtime errors during the backpropagation. The results are reported in~Table~\ref{tab:essential-stability}. Our method is $100\%$ stable compared to the baseline, which is $30\%$ stable, and it is also $\times 10$ faster. $\nabla$-RANSAC$^{*}$ denotes the implementation of the original $\nabla$-RANSAC without dropping non-real solutions during the forward pass of the solver since to backpropagate the gradients with the DDN, we need to find at least one solution to the optimization problem, so~in~case the~solver can~not find any real solution, we return the one from the previous iteration. We also measured the stability of~the~$\nabla$-RANSAC$^{*}$ to be sure that this minor change does~not contribute to~the~stability of our method.

All the experiments were conducted on NVIDIA GeForce GTX 1080 Ti with CUDA 12.2 and Pytorch 1.12.1 with Adam optimizer~\cite{DBLP:journals/corr/KingmaB14}, learning rate $10^{-4}$ and batch size $32$.

\section{Conclusion}\label{sec:Conclusion}
We have presented a new practical approach to backpropagating through minimal problem solvers. Our MinBackProp allows backpropagation either using the~Deep Declarative Networks machinery, which brings stability and easy use, or using the Implicit function theorem directly, which on top of stability, also speeds up significantly the backward pass. 
Through synthetic examples, we have shown the applicability of our method on a wide variety of tasks. Furthermore, we compared MinBackProp against the state-of-the-art relative pose estimation approach from~\cite{Wei_2023_ICCV} and have shown 100\% stability compared to $70-80\%$ failure rate of the autograd approach, while speeding up the computation $10$ times. Our method opens up a promising direction for efficient backpropagation through hard minimal problems. 

See Supplementary Material for additional technical details. The code is available at \url{https://github.com/disungatullina/MinBackProp} .

\section*{Acknowledgments}
This work has been supported by the EU H2020 No.\,871245 SPRING project.

\bibliographystyle{ieeenat_fullname}
\bibliography{all}

\begin{thebibliography}{36}
\providecommand{\natexlab}[1]{#1}
\providecommand{\url}[1]{\texttt{#1}}
\expandafter\ifx\csname urlstyle\endcsname\relax
  \providecommand{\doi}[1]{doi: #1}\else
  \providecommand{\doi}{doi: \begingroup \urlstyle{rm}\Url}\fi

\bibitem[Amos and Kolter(2017)]{DBLP:conf/icml/AmosK17}
Brandon Amos and J.~Zico Kolter.
\newblock Optnet: Differentiable optimization as a layer in neural networks.
\newblock In \emph{Proceedings of the 34th International Conference on Machine
  Learning, {ICML} 2017, Sydney, NSW, Australia, 6-11 August 2017}, pages
  136--145. {PMLR}, 2017.

\bibitem[Arandjelovi\'{c} and Zisserman(2012)]{6248018}
Relja Arandjelovi\'{c} and Andrew Zisserman.
\newblock Three things everyone should know to improve object retrieval.
\newblock In \emph{2012 IEEE Conference on Computer Vision and Pattern
  Recognition}, pages 2911--2918, 2012.

\bibitem[Barath et~al.(2020)Barath, Noskova, Ivashechkin, and
  Matas]{barath2019magsacplusplus}
Daniel Barath, Jana Noskova, Maksym Ivashechkin, and Jiri Matas.
\newblock {MAGSAC}++, a fast, reliable and accurate robust estimator.
\newblock In \emph{Conference on Computer Vision and Pattern Recognition},
  2020.

\bibitem[Bengio et~al.(1994)Bengio, Simard, and
  Frasconi]{DBLP:journals/tnn/BengioSF94}
Yoshua Bengio, Patrice~Y. Simard, and Paolo Frasconi.
\newblock Learning long-term dependencies with gradient descent is difficult.
\newblock \emph{{IEEE} Trans. Neural Networks}, 5\penalty0 (2):\penalty0
  157--166, 1994.

\bibitem[Brachmann and Rother(2019)]{Brachmann:2019}
Eric Brachmann and Carsten Rother.
\newblock Neural-guided ransac: Learning where to sample model hypotheses.
\newblock pages 4321--4330, 2019.

\bibitem[Brachmann et~al.(2017)Brachmann, Krull, Nowozin, Shotton, Michel,
  Gumhold, and Rother]{Brachmann2017}
Eric Brachmann, Alexander Krull, Sebastian Nowozin, Jamie Shotton, Frank
  Michel, Stefan Gumhold, and Carsten Rother.
\newblock Dsac~-- differentiable ransac for camera localization.
\newblock pages 2492--2500, 2017.

\bibitem[Chen et~al.(2020)Chen, Parra, Cao, Li, and
  Chin]{DBLP:conf/cvpr/0009PCLC20}
Bo Chen, {\'{A}}lvaro Parra, Jiewei Cao, Nan Li, and Tat{-}Jun Chin.
\newblock End-to-end learnable geometric vision by backpropagating pnp
  optimization.
\newblock In \emph{2020 {IEEE/CVF} Conference on Computer Vision and Pattern
  Recognition, {CVPR} 2020, Seattle, WA, USA, June 13-19, 2020}, pages
  8097--8106. Computer Vision Foundation / {IEEE}, 2020.

\bibitem[Chen et~al.(2022)Chen, Wang, Wang, Tian, Xiong, and
  Li]{DBLP:conf/cvpr/ChenWWTX022}
Hansheng Chen, Pichao Wang, Fan Wang, Wei Tian, Lu Xiong, and Hao Li.
\newblock Epro-pnp: Generalized end-to-end probabilistic perspective-n-points
  for monocular object pose estimation.
\newblock In \emph{{IEEE/CVF} Conference on Computer Vision and Pattern
  Recognition, {CVPR} 2022, New Orleans, LA, USA, June 18-24, 2022}, pages
  2771--2780. {IEEE}, 2022.

\bibitem[Cox et~al.(1998)Cox, Little, and O'Shea]{Cox-UAG-1998}
David Cox, John Little, and Donald O'Shea.
\newblock \emph{Using Algebraic Geometry}.
\newblock Springer, 1998.

\bibitem[Cox et~al.(2015)Cox, Little, and O'Shea]{Cox-IVA-2015}
David~A. Cox, John Little, and Donald O'Shea.
\newblock \emph{Ideals, Varieties, and Algorithms: An Introduction to
  Computational Algebraic Geometry and Commutative Algebra}.
\newblock Springer, 2015.

\bibitem[Cybernet Systems~Co.()]{Maple}
Ltd. Cybernet Systems~Co.
\newblock Maple.
\newblock http://www.maplesoft.com/products/maple/.

\bibitem[Dang et~al.(2018)Dang, Yi, Hu, Wang, Fua, and
  Salzmann]{DBLP:conf/eccv/DangYHWFS18}
Zheng Dang, Kwang~Moo Yi, Yinlin Hu, Fei Wang, Pascal Fua, and Mathieu
  Salzmann.
\newblock Eigendecomposition-free training of deep networks with zero
  eigenvalue-based losses.
\newblock In \emph{Computer Vision - {ECCV} 2018 - 15th European Conference,
  Munich, Germany, September 8-14, 2018, Proceedings, Part {V}}, pages
  792--807. Springer, 2018.

\bibitem[Elkadi and Mourrain(2005)]{Elkadi2005}
Mohamed Elkadi and Bernard Mourrain.
\newblock \emph{Symbolic-numeric methods for solving polynomial equations and
  applications}, pages 125--168.
\newblock Springer Berlin Heidelberg, Berlin, Heidelberg, 2005.

\bibitem[Fischler and Bolles(1981)]{DBLP:journals/cacm/FischlerB81}
Martin~A. Fischler and Robert~C. Bolles.
\newblock Random sample consensus: {A} paradigm for model fitting with
  applications to image analysis and automated cartography.
\newblock \emph{Commun. {ACM}}, 24\penalty0 (6):\penalty0 381--395, 1981.

\bibitem[Gould et~al.(2022)Gould, Hartley, and Campbell]{Gould2022}
Stephen Gould, Richard Hartley, and Dylan Campbell.
\newblock Deep declarative networks.
\newblock \emph{IEEE Transactions on Pattern Analysis and Machine
  Intelligence}, 44\penalty0 (8):\penalty0 3988--4004, 2022.

\bibitem[Hartley and Zisserman(2003)]{HZ-2003}
Richard Hartley and Andrew Zisserman.
\newblock \emph{Multiple View Geometry in Computer Vision}.
\newblock Cambridge, 2nd edition, 2003.

\bibitem[Hartley(1995)]{10.5555/839277.840009}
R.~I. Hartley.
\newblock In defence of the 8-point algorithm.
\newblock In \emph{Proceedings of the Fifth International Conference on
  Computer Vision}, page 1064, USA, 1995. IEEE Computer Society.

\bibitem[HASHEMI and LAZARD(2011)]{doi:10.1142/S0218196711006364}
AMIR HASHEMI and DANIEL LAZARD.
\newblock Sharper complexity bounds for zero-dimensional groebner bases and
  polynomial system solving.
\newblock \emph{International Journal of Algebra and Computation}, 21\penalty0
  (05):\penalty0 703--713, 2011.

\bibitem[Ionescu et~al.(2015)Ionescu, Vantzos, and Sminchisescu]{Ionescu2015}
Catalin Ionescu, Orestis Vantzos, and Cristian Sminchisescu.
\newblock Matrix backpropagation for deep networks with structured layers.
\newblock In \emph{2015 IEEE International Conference on Computer Vision
  (ICCV)}, pages 2965--2973, 2015.

\bibitem[Jin et~al.(2020)Jin, Mishkin, Mishchuk, Matas, Fua, Yi, and
  Trulls]{PhotoTourism2020}
Yuhe Jin, Dmytro Mishkin, Anastasiia Mishchuk, Jiri Matas, Pascal Fua,
  Kwang~Moo Yi, and Eduard Trulls.
\newblock Image matching across wide baselines: From paper to practice.
\newblock \emph{International Journal of Computer Vision}, 2020.

\bibitem[Kabsch(1976)]{Kabsch:a12999}
W. Kabsch.
\newblock {A solution for the best rotation to relate two sets of vectors}.
\newblock \emph{Acta Crystallographica Section A}, 32\penalty0 (5):\penalty0
  922--923, 1976.

\bibitem[Kingma and Ba(2015)]{DBLP:journals/corr/KingmaB14}
Diederik~P. Kingma and Jimmy Ba.
\newblock Adam: {A} method for stochastic optimization.
\newblock In \emph{3rd International Conference on Learning Representations,
  {ICLR} 2015, San Diego, CA, USA, May 7-9, 2015, Conference Track
  Proceedings}, 2015.

\bibitem[Kukelova et~al.(2008)Kukelova, Bujnak, and
  Pajdla]{kukelova2008automatic}
Zuzana Kukelova, Martin Bujnak, and Tomas Pajdla.
\newblock Automatic generator of minimal problem solvers.
\newblock In \emph{ECCV}, 2008.

\bibitem[Larsson et~al.(2017)Larsson, Kukelova, and Zheng]{larsson2017making}
Viktor Larsson, Zuzana Kukelova, and Yinqiang Zheng.
\newblock Making minimal solvers for absolute pose estimation compact and
  robust.
\newblock In \emph{ICCV}, 2017.

\bibitem[Larsson et~al.(2018)Larsson, Oskarsson, {\AA}str{\"{o}}m, Wallis,
  Kukelova, and Pajdla]{Larsson-CVPR-2018}
Viktor Larsson, Magnus Oskarsson, Kalle {\AA}str{\"{o}}m, Alge Wallis, Zuzana
  Kukelova, and Tom{\'{a}}s Pajdla.
\newblock Beyond grobner bases: Basis selection for minimal solvers.
\newblock In \emph{2018 {IEEE} Conference on Computer Vision and Pattern
  Recognition, {CVPR} 2018, Salt Lake City, UT, USA, June 18-22, 2018}, pages
  3945--3954, 2018.

\bibitem[Martyushev et~al.(2022)Martyushev, Vr{\'{a}}bl{\'{\i}}kov{\'{a}}, and
  Pajdla]{DBLP:conf/cvpr/MartyushevVP22}
Evgeniy Martyushev, Jana Vr{\'{a}}bl{\'{\i}}kov{\'{a}}, and Tom{\'{a}}s Pajdla.
\newblock Optimizing elimination templates by greedy parameter search.
\newblock In \emph{{IEEE/CVF} Conference on Computer Vision and Pattern
  Recognition, {CVPR} 2022, New Orleans, LA, USA, June 18-24, 2022}, pages
  15733--15743. {IEEE}, 2022.

\bibitem[Meyer(2001)]{Meyer-2001}
Carl~D. Meyer.
\newblock \emph{Matrix Analysis and Applied Linear Algebra}.
\newblock {SIAM}: Society for Industrial and Applied Mathematics, Philadelphia,
  PA, USA, 2001.

\bibitem[Nist{\'e}r(2004)]{Nistr2004AnES}
David Nist{\'e}r.
\newblock An efficient solution to the five-point relative pose problem.
\newblock \emph{IEEE Transactions on Pattern Analysis and Machine
  Intelligence}, 26:\penalty0 756--770, 2004.

\bibitem[Ranftl and Koltun(2018)]{Ranftl2018}
Rene Ranftl and Vladlen Koltun.
\newblock Deep fundamental matrix estimation.
\newblock In \emph{The European Conference on Computer Vision (ECCV)}, 2018.

\bibitem[Rudin(1976)]{Rudin-1976}
Walter Rudin.
\newblock \emph{Principles of Mathematical Analysis}.
\newblock McGraw-Hill, 1976.

\bibitem[Shafarevich and Reid(2013)]{shafarevich2013basic}
I.R. Shafarevich and M. Reid.
\newblock \emph{Basic Algebraic Geometry 1: Varieties in Projective Space}.
\newblock Springer Berlin Heidelberg, 2013.

\bibitem[Stew{\'{e}}nius et~al.(2005)Stew{\'{e}}nius, Nist{\'{e}}r, Kahl, and
  Schaffalitzky]{DBLP:conf/cvpr/SteweniusNKS05}
Henrik Stew{\'{e}}nius, David Nist{\'{e}}r, Fredrik Kahl, and Frederik
  Schaffalitzky.
\newblock A minimal solution for relative pose with unknown focal length.
\newblock In \emph{2005 {IEEE} Computer Society Conference on Computer Vision
  and Pattern Recognition {(CVPR} 2005), 20-26 June 2005, San Diego, CA,
  {USA}}, pages 789--794. {IEEE} Computer Society, 2005.

\bibitem[Wahba(1965)]{Wahba1965}
Grace Wahba.
\newblock A least squares estimate of satellite attitude.
\newblock \emph{SIAM Review}, 7\penalty0 (3):\penalty0 409--409, 1965.

\bibitem[Wei et~al.(2023)Wei, Patel, Shekhovtsov, Matas, and
  Barath]{Wei_2023_ICCV}
Tong Wei, Yash Patel, Alexander Shekhovtsov, Jiri Matas, and Daniel Barath.
\newblock Generalized differentiable ransac.
\newblock In \emph{Proceedings of the IEEE/CVF International Conference on
  Computer Vision (ICCV)}, pages 17649--17660, 2023.

\bibitem[Zhang et~al.(2020)Zhang, Gu, Michalkiewicz, Baktashmotlagh, and
  Eriksson]{DBLP:journals/corr/abs-2003-01822}
Qianggong Zhang, Yanyang Gu, Mateusz Michalkiewicz, Mahsa Baktashmotlagh, and
  Anders~P. Eriksson.
\newblock Implicitly defined layers in neural networks.
\newblock \emph{CoRR}, abs/2003.01822, 2020.

\bibitem[Zhao et~al.(2021)Zhao, Ge, Zhu, Zhao, Li, and
  Salzmann]{zhao2021progressive}
C. Zhao, Y. Ge, F. Zhu, R. Zhao, H. Li, and M. Salzmann.
\newblock Progressive correspondence pruning by consensus learning.
\newblock In \emph{2021 IEEE/CVF International Conference on Computer Vision
  (ICCV)}, pages 6444--6453, Los Alamitos, CA, USA, 2021. IEEE Computer
  Society.

\end{thebibliography}

\clearpage
\newpage
\clearpage
\setcounter{page}{1}

\begin{strip}
\centering
\Large{SUPPLEMENTARY MATERIAL}
\vspace{0.5cm}
\end{strip}

\section{Details on using the Implicit function theorem and DDN to backpropagate through minimal problem solvers}
\label{sec:dx/da}
In this section,  we give additional details on using the~Implicit function theorem and implementing backpropagation through minimal solvers by the DDN framework.
\subsection{Proof of~\cref{thm:dx/da}~(Differentiating roots of a polynomial system)}
Let us proof~\cref{{thm:dx/da}}: 

\noindent\usebox{\thmdxda}

The proof is obtained by adapting the standard Implicit function theorem~\cite{Rudin-1976} to polynomials.
\begin{theorem}[Implicit function theorem]
    Let $f({\bf x},{\bf a})$ be a continuously differentiable mapping of an open set $E \subseteq \RR^{N+M}$ into $\RR^N$ such that $f({\bf y},{\bf b}) = 0$ for some point $({\bf y},{\bf b}) \in E$. Assume that $\left[\frac{\partial f({\bf x},{\bf a})}{\partial {\bf x}}({\bf y},{\bf b})\right]$ is invertible. 
    
    Then, there exist open sets $U \subseteq \RR^{N+M}$ and $W \subseteq \RR^M$, with $({\bf y},{\bf b}) \in U$ and ${\bf b} \in W$, having the following property: To every ${\bf a} \in W$ corresponds a unique ${\bf x}$ such that $({\bf x},{\bf a}) \in U$ and $f({\bf x},{\bf a}) = 0$. 

    If this ${\bf x}$ is defined to be $g({\bf a})$, then $g$ is a continuously differentiable mapping of $W$ into $\RR^N$ with ${\bf y} = g({\bf b})$, $f(g({\bf a}),{\bf a})=0$, $\forall {\bf a}\in W$, and
    {\small
    \[
    \left[\frac{\partial g({\bf a})}{\partial {\bf a}}({\bf b})\right] \!\!=\!
    - \left[\frac{\partial f({\bf x},{\bf a})}{\partial {\bf x}}({\bf y},{\bf b})\right]^{\!-1}\!\!
    \left[\frac{\partial f({\bf x},{\bf a})}{\partial {\bf a}}({\bf y},{\bf b})\right] \text{.}
    \]
    }
\label{thm:IFT}
\end{theorem}

\vspace*{-2em}
\begin{proof} See~\cite[{\em pp}.\,224--227]{Rudin-1976}. \end{proof}
\begin{proof}(of \cref{thm:dx/da})
\cref{thm:dx/da} is an adaptation of~\cref{thm:IFT}. First, we slightly change the notation for Jacobians to stress the explicit indexing, explicitly name the~components of $f$ as $h_i$, and use ${\bf x}({\bf a})$ for $g({\bf a})$. We also consider the map $f$ to consist of $K \geq N$ polynomials $h_k(x_1,\ldots,x_N), k=1,\ldots,K$. Assuming $K \geq N$ corresponds to having polynomial systems with more equations than unknowns to allow for general polynomial systems. Of~course, if~the~Jacobian of such $f$ w.r.t.\ ${\bf x}$ has full rank at ${\bf b}$, we can choose a map $f_N$ of $N$ polynomials from all $K$ polynomials with invertible Jacobian at~${\bf b}$. Alternatively, we can use the pseudoinverse for the~Jacobian of full $f$.

Now, let us show that the assumption of~\cref{thm:dx/da} are for polynomials equivalent to the assumptions of~\cref{thm:IFT}. First, consider that polynomials are continuously differentiable on whole $\RR^N$ and hence we only need to show that the Jacobian $\left[\frac{\partial h_k({\bf x},{\bf a})}{\partial x_n}({\bf x}({\bf b}),{\bf b})\right]$ has full rank if and only if ${\bf x}({\bf b})$ is an isolated multiplicity-one solution. This is a standard fact from algebraic geometry. 

The reasoning is, in general, as follows: First, the dimension of an irreducible variety (over $\CC$) is given by the rank of the Jacobian of its vanishing ideal at a~generic point~\cite[Ch.~2]{shafarevich2013basic}.  Next, for a square polynomial system $F$, if the Jacobian of $F$ has full rank at a point, then this is an upper bound on the rank of the Jacobian of the vanishing ideal of that point, and hence the point is isolated. Finally, a proof of the~"multiplicity-one" characterization depends on how one defines multiplicity. Taking the approach of~\cite{Cox-UAG-1998}, one can use the dimension of the local ring at an isolated point. With that definition, it is straightforward to prove that multiplicity-one appears exactly when the Jacobian is nonsingular.
\end{proof}

\subsection{Justification of implementing backpropagation through minimal problem solvers by DDN}

Let us present additional details on implementing the backpropagation through minimal problem solvers by the DNN framework. The main point for justifying the~use of DDN is to show that the loss function and constraints used in the DDN framework indeed provide the solution to a minimal problem solver. 

In DDN, we formulate a minimal problem solver as minimizing the sum of squares of data-dependent polynomial equations $r(\mathrm{w}, \mathrm{y})$ subject to data-independent polynomial constraints $h_i(\mathrm{y})$. 
\begin{equation*}
\begin{aligned}
 &\mathrm{\hat y} (\mathrm{w}) \in \underset{\mathrm{y \text{\,} \in \text{\,} \mathrm{C}}}{\argmin}\,
    \sum_{m=1}^M(r_m(\mathrm{w}, \mathrm{y}))^2 \\
    & \hspace{1.5cm} \text{s.t. } 
    h_i(\mathrm{y})=0\text{ .}
\end{aligned}
\end{equation*}
We know that such a loss can reach zero in finitely many points while satisfying the constraints. This is a particularly meaningful formulation when the data-dependent equations are linear, but other formulations might be useful in other situations. This gives lagrangian 
\begin{equation*}
    L(\mathrm{y},\lambda) = (r(\mathrm{w}, \mathrm{y}))^2 + \sum_{i=1}^K \lambda_i h_i(\mathrm{y})
\end{equation*}
and thus, we are getting equations
\begin{equation*}
\begin{aligned}
    0 = \frac{\partial L(\mathrm{y},\lambda)}{\partial \mathrm{y}} =
    2\,\sum_{m=1}^M r_m(\mathrm{w}, \mathrm{y})\frac{\partial r_m(\mathrm{w}, \mathrm{y})}{\partial \mathrm{y}} 
    + \sum_{i=1}^K \lambda_i \frac{\partial h_i(\mathrm{y})}{\partial \mathrm{y}}
    \\
    0 = \frac{\partial L(\mathrm{y},\lambda)}{\partial \lambda} = h_i(\mathrm{y}) \hspace{0.655\linewidth}
\end{aligned}    
\end{equation*}
We see that the above equations are satisfied for the solutions of the original minimal problem $r_m(\mathrm{w},\mathrm{y})=0$, $h_i(\mathrm{y})=0$. Additional solutions appear when the Jacobian is not full rank or some $\lambda_i$ are non-zero, but these are never a problem in our approach. First, in the forward pass, we compute only the solutions to the original minimal problem by a tailored algebraic minimal problem solver. Secondly, in the backward pass, we use the~DDN formulation locally only, and thus, additional solutions do not affect the values of the derivatives at the solution of the minimal problem; hence using the~DDN correctly backpropagates through the minimal problems.

\begin{table*}[h]
\centering
\resizebox{\textwidth}{!}{
\begin{tabular}{|c|ccc|ccc|ccc|}
\hline Scene / Method \& metric & \makecell{$\nabla$-RANSAC~\cite{Wei_2023_ICCV}\\AUC@$5^{\circ}$} & \makecell{MinBackProp DDN\\AUC@$5^{\circ}$} & \makecell{MinBackProp IFT\\AUC@$5^{\circ}$} & \makecell{$\nabla$-RANSAC~\cite{Wei_2023_ICCV}\\AUC@$10^{\circ}$} & \makecell{MinBackProp DDN\\AUC@$10^{\circ}$} & \makecell{MinBackProp IFT\\AUC@$10^{\circ}$} & \makecell{$\nabla$-RANSAC~\cite{Wei_2023_ICCV}\\AUC@$20^{\circ}$} & \makecell{MinBackProp DDN\\AUC@$20^{\circ}$} & \makecell{MinBackProp IFT\\AUC@$20^{\circ}$} \\
\hline Buckingham Palace & 0.23 & 0.23 & 0.24 & 0.28 & 0.28 & 0.28 & 0.34 & 0.33 & 0.34 \\
Brandenburg Gate & 0.51 & 0.49 & 0.52 & 0.57 & 0.55 & 0.58 & 0.64 & 0.61 & 0.65 \\
Colosseum Exterior & 0.31 & 0.31 & 0.31 & 0.33 & 0.33 & 0.33 & 0.35 & 0.36 & 0.35 \\
Grand Place Brussels & 0.17 & 0.16 & 0.18 & 0.22 & 0.21 & 0.23 & 0.27 & 0.26 & 0.29 \\
Notre Dame Front Facade & 0.42 & 0.43 & 0.42 & 0.47 & 0.47 & 0.47 & 0.52 & 0.51 & 0.51 \\
Palace of Westminster & 0.36 & 0.35 & 0.39 & 0.4 & 0.39 & 0.43 & 0.44 & 0.44 & 0.47 \\
Pantheon Exterior & 0.61 & 0.59 & 0.61 & 0.67 & 0.64 & 0.67 & 0.73 & 0.71 & 0.73 \\
Prague Old Town Square  & 0.17 & 0.17 & 0.16 & 0.19 & 0.19 & 0.18 & 0.21 & 0.22 & 0.21 \\
Sacre Coeur  & 0.74 & 0.72 & 0.73 & 0.75 & 0.74 & 0.75 & 0.77 & 0.76 & 0.77 \\
Taj Mahal  & 0.57 & 0.58 & 0.59 & 0.61 & 0.62 & 0.63 & 0.66 & 0.66 & 0.67 \\
Trevi Fountain  & 0.33 & 0.33 & 0.33 & 0.37 & 0.36 & 0.37 & 0.41 & 0.4 & 0.42 \\
Westminster Abbey  & 0.44 & 0.45 & 0.45 & 0.51 & 0.52 & 0.52 & 0.57 & 0.58 & 0.58 \\
\hline Avg. over all scenes & 0.405 & 0.401 & \textbf{0.411} & 0.448 & 0.442 & \textbf{0.453} & 0.493 & 0.487 & \textbf{0.499}\\
\hline
\end{tabular}}
\caption{The AUC scores for essential matrix estimation of $\nabla$-RANSAC~\cite{Wei_2023_ICCV} and MinBackProp (ours) for each of~the~12~scenes of~the~PhotoTourism dataset~\cite{PhotoTourism2020}, under different thresholds, described in \cref{sec:Exp:EG}. Our method shows the same result as $\nabla$-RANSAC.}
\label{tab:scenes}
\end{table*}

\section{SVD}
\label{supp:fundamental}
In this section, we define a formulation for the toy example for fundamental matrix estimation mentioned in~\cref{sec:Exp:3D}.
\subsection{Fundamental matrix estimation with an outlier}
Let us consider a toy example for fundamental matrix estimation with an outlier. Given a set of tentative correspondences $\p_i \leftrightarrow \tilde{\p}_i$, $\p_i \in I_1$ and $\tilde{\p}_i \in I_2$, where $I_1$ and $I_2$ are two synthetic images with different camera poses of one synthetic scene, $i = \overline{1, N}$, and $N$ is the number of correspondences. We want to estimate the fundamental matrix $\F \in \mathbb{R}^{3 \times 3}$: $\tilde{l}_i = \F \p_i$, $l_i = \F^T \tilde{\p}_i$, where $l_i$ and $\tilde{l}_i$ are epipolar lines on the $I_1$ and $I_2$, respectively. Following~\cref{sec:Exp:3D}, given the ground-truth fundamental matrix $\mathrm{F_{true}} \in \mathbb{R}^{3 \times 3}$, we define the weights $\mathrm{w} = [w_1, \dots w_N]$ for each correspondence to~mark inliers and~outliers. 

\begin{figure}
    \centering
     \includegraphics[width=1.05\linewidth]{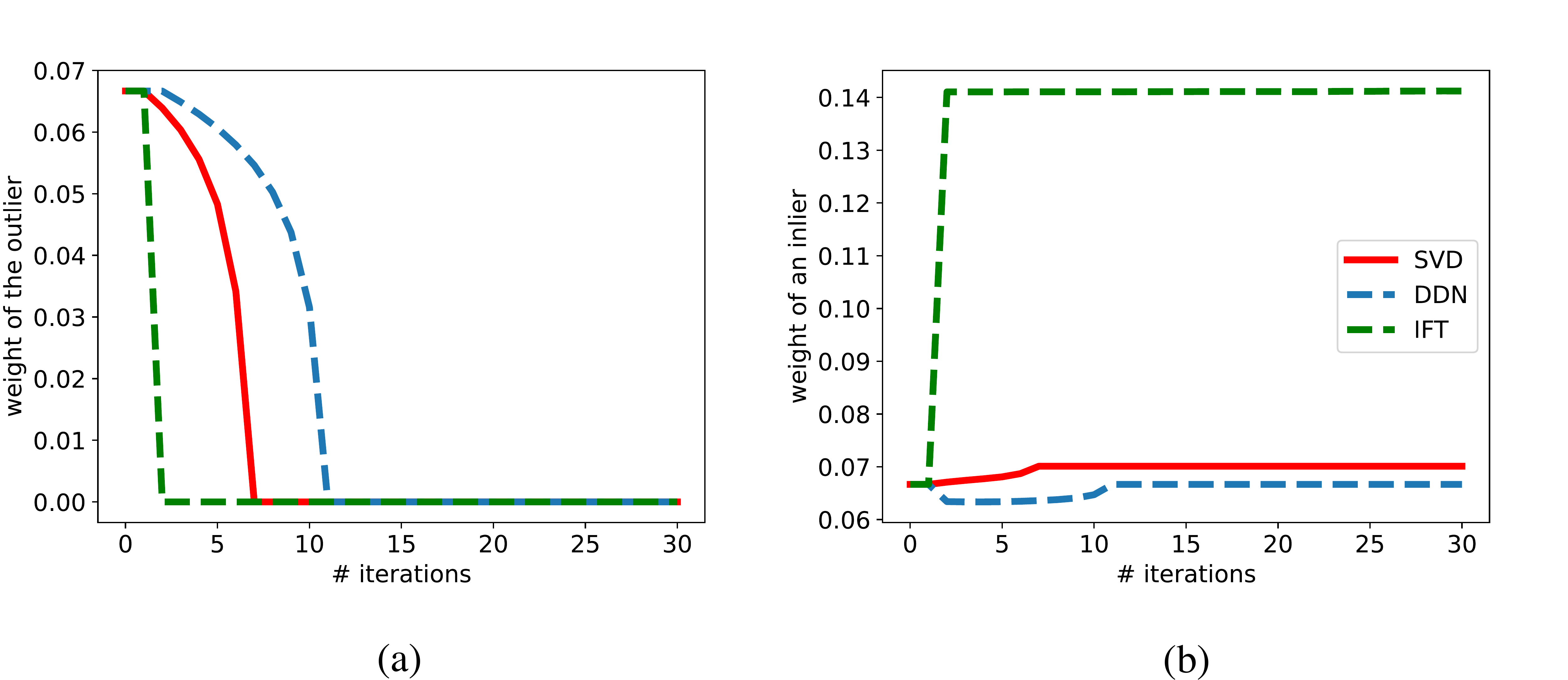}
    \caption{A toy example for the fundamental matrix estimation with an outlier: (a) the weight for the outlier and (b)
the weight for an inlier during the optimization. The red indicates explicit computation of the gradients, and the green/blue denotes backpropagation with our MinBackProp. One can see that the gradients computed with the Implicit Function theorem and the Deep Declarative Networks within our least squares formulation with constraints~\cref{eq:supp:fundamental} approximate the gradients of SVD computed in closed form.}
    \label{fig:supp:fund_in_out}
\end{figure}

To find $\F$ and $\mathrm{w}$ within our MinBackProp framework, we define a~bi-level optimization problem as~\cref{eq:formulation}:
\begin{equation}
\begin{aligned}
\underset{\mathrm{w}}{\min}\,& \left\| \Fh (\mathrm{w})  - \mathrm{F_{true}} \right\|_2^2  \\
\text{s.\,t. } &\Fh (\mathrm{w}) \in \underset{\text{\,} \F \in \text{\,} \mathbb{R}^{3 \times 3}}{\argmin}\,
    \frac{1}{N} \sum_{i=1}^{N} w_i \text{\,} \left( \tilde{\p}_i^T \F \p_i \right)^2 \\
    & \hspace{1.5cm} \text{s.\,t. }  \det(\F) = 0 \text{ ,} \\
    & \hspace{2.56cm}   \left\| \F \right\|^2 = 1 \text{ .}
\end{aligned}
\label{eq:supp:fundamental}
\end{equation}
In our experiment, $N = 15$, with $14$ inliers and one outlier, $\mathrm{w}$ is initialized uniformly with~$1/N$. We optimize the weights with gradient descent with a learning rate of $1000$ for $30$ iterations. We use the  8-point algorithm~\cite{10.5555/839277.840009} (SVD) to solve the low-level optimization problem. As in~\cref{sec:Exp:3D}, forward pass is~the~same for~both explicit and implicit methods. For~the~backward pass, we both compute the gradients of SVD in closed form~\cite{Ionescu2015} and using the Implicit Function theorem and the Deep Declarative Networks~\cite{Gould2022}. The~result of the~optimization for the~outlier's and an inlier's weights is shown in~Figure~\ref{fig:supp:fund_in_out}. Our DDN/IFT gradients approximate the gradients of SVD in closed form.


\section{Details on the implementation of~the~MinBackProp IFT} \label{supp:essential}
Here we present the system of polynomial equations to~compute the derivatives for the MinBackProp IFT in \cref{sec:Exp:EG}
\begin{equation}
H(\E, \q_i, \tilde{\q}_i) = 
    \begin{cases}
    & \tilde{\q}_1^T \E \q_1  = 0 \text{ ,}\\
    & \tilde{\q}_2^T \E \q_2  = 0 \text{ ,}\\
    & \tilde{\q}_3^T \E \q_3  = 0 \text{ ,}\\
    & \tilde{\q}_4^T \E \q_4  = 0 \text{ ,}\\
    & \tilde{\q}_5^T \E \q_5  = 0 \text{ ,}\\
    &  \left\| \E \right\|^2 = 1 \text{ ,}\\
    & 2\E\E^T\E - \tr \left( \E\E^T \right)\E = 0 \text{ ;}\\ 
    \end{cases}
\end{equation}
where $w_i = [\q_i,\text{\,}\tilde{\q}_i]$ is a pair of matches normalized by the intrinsic matrix pixel coordinates, $\E$ is the essential matrix, and $\left\| \cdot \right\|$ is the Frobenius norm of the matrix.

To compute the Jacobian $\frac{\partial \hat{E}}{\partial w}$, according to the Implicit function theorem, we need to compute the Jacobians $\frac{\partial H}{\partial \E}$ and $\frac{\partial H}{\partial w}$, and to invert the Jacobian $\frac{\partial H}{\partial \E}$. To accomplish the~inversion, we choose the first six equations of $H$ and construct the last three as a linear combination of the last nine equations of $H$. To check if the matrix is full-rank we use SVD. If~the~rank is lower than nine, one can randomly choose a~full-rank Jacobian from the~same batch to invert.

\end{document}